%% file: 00_main.tex
\documentclass[letterpaper, 10 pt, conference]{ieeeconf}  

\IEEEoverridecommandlockouts                              
\usepackage{amsmath}
\usepackage{mathtools}
\usepackage{cite}
\usepackage[ruled, lined, linesnumbered, commentsnumbered, longend, noend]{algorithm2e}
\usepackage[font=small, labelfont=bf]{caption}
\usepackage{hyperref}
\usepackage{amsmath,amssymb,amsfonts}
\usepackage{algorithmic}
\usepackage{graphicx}
\usepackage{textcomp}
\usepackage{xcolor}
\usepackage{breqn}
\usepackage{gensymb}

\usepackage{fancyhdr} 
\def\BibTeX{{\rm B\kern-.05em{\sc i\kern-.025em b}\kern-.08em
    T\kern-.1667em\lower.7ex\hbox{E}\kern-.125emX}}
 
\newcommand{\argminE}{\mathop{\mathrm{argmin}}}

\definecolor{pink}{RGB}{255, 192, 203}

\title{\LARGE \bf Real-Time 3D Semantic Scene Perception for \\Egocentric Robots with Binocular Vision}

\author{\large Khang Nguyen \hspace{45pt} Tuan Dang \hspace{45pt} Manfred Huber {\footnotesize \thanks{All authors are with the Learning and Adaptive Robotics Laboratory, Department of Computer Science and Engineering, University of Texas at Arlington, Arlington, TX 76013, USA. (emails: \href{mailto:khang.nguyen8@mavs.uta.edu}{\text{khang.nguyen8@mavs.uta.edu}}, \href{mailto:tuan.dang@uta.edu}{\text{tuan.dang@uta.edu}}, \href{mailto:huber@cse.uta.edu}{\text{huber@cse.uta.edu}})}}}

\begin{document}

\maketitle
\thispagestyle{fancy}

\begin{abstract}
Perceiving a three-dimensional (3D) scene with multiple objects while moving indoors is essential for vision-based mobile cobots, especially for enhancing their manipulation tasks. In this work, we present an end-to-end pipeline with instance segmentation, feature matching, and point-set registration for egocentric robots with binocular vision and demonstrate the robot's grasping capability through the proposed pipeline. First, we design an RGB image-based segmentation approach for single-view 3D semantic scene segmentation, leveraging common object classes in 2D datasets to encapsulate 3D points into point clouds of object instances through corresponding depth maps. Next, 3D correspondences of two consecutive segmented point clouds are extracted based on matched keypoints between objects of interest in RGB images from the prior step. In addition, to be aware of spatial changes in 3D feature distribution, we also weigh each 3D point pair based on the estimated distribution using kernel density estimation (KDE), which subsequently gives robustness with less central correspondences while solving for rigid transformations between point clouds. Finally, we test our proposed pipeline on the 7-DOF dual-arm Baxter robot with a mounted Intel RealSense D435i RGB-D camera. The result shows that our robot can segment objects of interest, register multiple views while moving, and grasp the target object. The source code is available at https://github.com/mkhangg/semantic\_scene\_perception.
\end{abstract}

\input{01_introduction}
\input{02_related_work}
\input{03_segmentation}
\input{04_feature_matching}
\input{05_reconstruction}
\input{06_evaluation}
\input{07_limitations}
\input{08_conclusions}
\input{09_acknowledgement}

\bibliographystyle{IEEEtran}
\bibliography{IEEEabrv, 10_references}

\end{document}

%% file: 01_introduction.tex
\section{Introduction}
Egocentric vision is crucial for machine and human vision, especially in a dense environment, due to its high selective attention to objects of interest while ignoring non-relevant entities (\textit{e.g.}, walls, floor, etc.) \cite{smith2018developing, humblot2022navigation}. From the standpoint of 3D perception for autonomous robots, the spatial information of objects of interest is also needed to ameliorate their manipulation tasks. Indeed, currently segmentation and registration tasks are often done separately \textit{from the point cloud perspective} while considering every 3D point presented in the scene. Yet, deploying both procedures simultaneously may result in expensive computations on resource-constraint robots. For this reason, enabling a lightweight egocentric 3D segmentation, feature matching, and scene reconstruction pipeline is essential for vision-based indoor mobile cobots. 

\begin{figure}[t]
    \centering
    \includegraphics[width=0.89\linewidth]{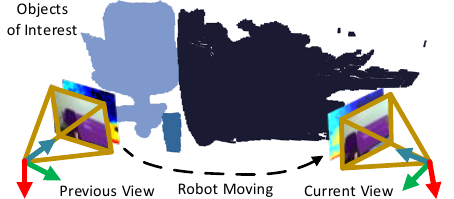}
    \vspace{-2pt}
    \caption{Overview of 3D semantic scene perception for vision-based indoor robots with binocular vision and multiple egocentric views.}
    \vspace{-19pt}
    \label{fig:segmatreg_concept}
\end{figure}

Considerable prior efforts are spent in learning matching features between images with deterministic algorithms \cite{lowe2004distinctive, bay2006surf, calonder2010brief, rublee2011orb} and with machine learning (ML) models \cite{detone2018superpoint, sarlin2020superglue}, which subsequently lead to robotics applications \cite{tomasi1992shape, mur2015orb, kelly2023target, lobefaro2023estimating}. These works made notable contributions to 3D scene perception; however, 3D semantic scene perception for indoor mobile cobots also needs spatial occupancy information of objects of interest, particularly in household settings, to initiate manipulation (\textit{e.g.}, appropriately grasping target objects).

\begin{figure*}[t]
    \centering
    \includegraphics[width=1.00\linewidth]{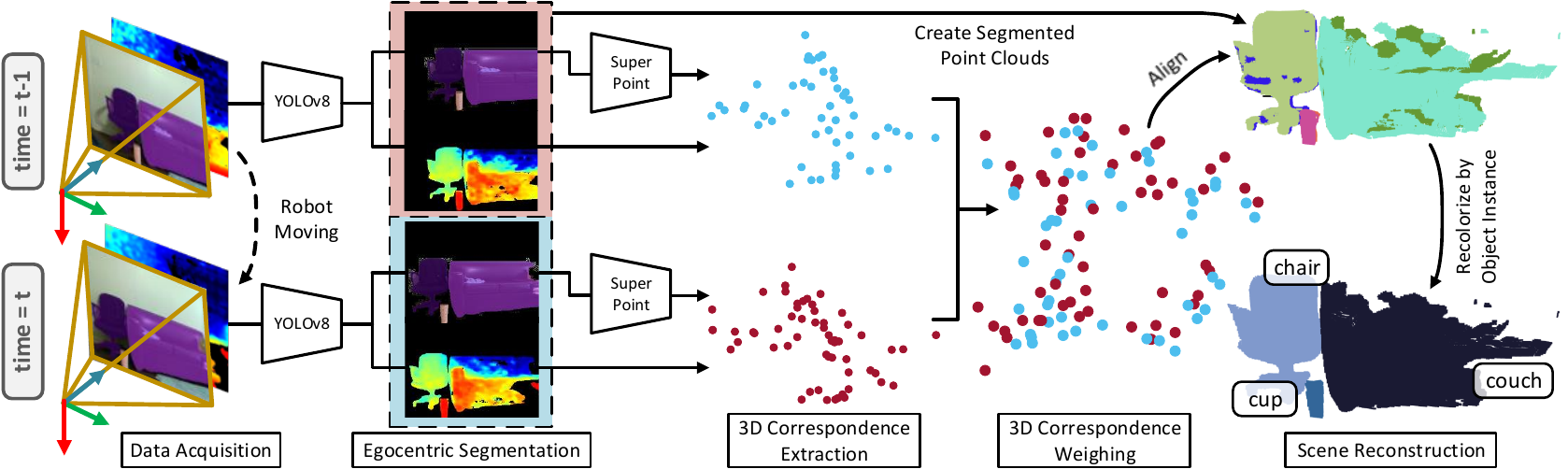}
    \vspace{-12pt}
    \caption{The 3D semantic scene perception pipeline when the robot takes two (or multiple) views of a scene includes (1) egocentric segmentation to create point clouds of objects of interest (Sec. \ref{sec:segmentation}), (2) extracting and matching corresponding features on masked RGB images to infer 3D correspondences via depth maps (Sec. \ref{sec:feature_matching}), (3) finding optimal transformations based on weighted 3D correspondences and reconstructing the 3D scene (Sec. \ref{sec:scene_reconstruction}), and (4) returning the aligned point cloud from multiple views with segmented objects.}
    \vspace{-17pt}
    \label{fig:method_overview}
\end{figure*}

Recently proposed methods mainly extract keypoints from images without focusing on objects of interest \cite{jau2020deep, tang2021self}, leading to expensive computation in later stages. Likewise, classical keypoint matching methods usually operate on entire images, but not regions of interest, which might result in mismatched or unwanted correspondences. Furthermore, the Iterative Closest Point-based (ICP) registration and ML-based methods for 3D multiview perform on a large set of 3D points without prior knowledge about the scene. 

Therefore, to fill gaps in previous works and improve 3D semantic scene perception for vision-based mobile cobots as illustrated in Fig. \ref{fig:segmatreg_concept}, we (1) use lightweight semantic segmentation, which focuses only on objects of interest, (2) map 2D correspondences between two RGB images into 3D correspondences between two point clouds and align them with the awareness of important points, and (3) design a complete pipeline for indoor mobile robots with binocular vision to achieve 3D semantic perception (Fig. \ref{fig:method_overview}) by using an RGB-D camera in real-time along with robot operations.

As performing these tasks on a robot without a dedicated Graphics Processing Unit (GPU) requires computational power and memory optimization, we face a few challenges when designing this pipeline. Indeed, performing deep learning-based (DL) for semantic segmentation and matching point sets requires a huge consideration and customization of the state-of-the-art models, which are also included in the pipeline. Also, reconstructing a scene from multiple views is a complex iterative task since the traditional registration methods usually perform on the entire point cloud inputs.

In this work, we address the aforementioned challenges and make the following contributions: (1) a robust method to extract and statistically weigh 3D correspondences for rigid point cloud alignment, (2) an end-to-end segmentation, feature matching, and global registration pipeline for egocentric robots with binocular vision, (3) tests with a real robot system to verify the correctness of our proposed method.

%% file: 02_related_work.tex
\section{Related Work}
\textbf{Egocentric 3D Object Segmentation}: 
Egocentric object segmentation has evolved rapidly in the past decade, from first-person view \cite{fathi2011learning, li2015delving, tschernezki2021neuraldiff} to robotics applications \cite{kundu20183d, li2020learning, humblot2022navigation}. Recently, this line of research has been enhanced into the domain of 3D computer vision, with DL models training on large computer-aided design (CAD) datasets \cite{kundu20183d, humblot2022navigation}. These works generally preclude themselves from segmenting objects that are unavailable in 3D datasets but present in 2D datasets. However, point cloud representations of objects, as well as scenes of multiple objects, in an egocentric view are composed of RGB and depth images from RGB-D cameras. Understanding these subcomponents and how point clouds are created, we leverage state-of-the-art segmentation models \cite{Jocher_YOLO_by_Ultralytics_2023} to segment objects represented in RGB images and thus encapsulate 3D points into point clouds of the recognized objects when fusing depth images during the point cloud acquisition process. Therefore, this design takes advantage of the richness of object classes in the MS COCO image dataset but is unavailable in CAD datasets to recognize objects from an egocentric view.

\begin{figure*}[t]
    \centering
    \includegraphics[width=1.00\linewidth]{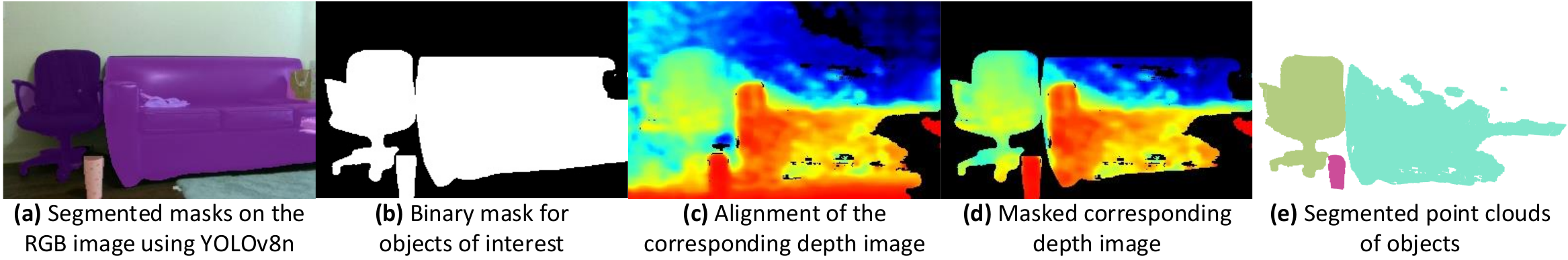}
    \vspace{-17pt}
    \caption{The egocentric object segmentation process includes \textbf{(a)} segmenting masks on the RGB image using YOLOv8n segmentation model, \textbf{(b)} obtaining and aggregating binary masks of the objects of interest, \textbf{(c)} aligning the corresponding depth image, \textbf{(d)} rectifying non-masked depth pixels on the aligned depth image with obtained masks, and \textbf{(e)} creating point clouds of such objects.}
    \vspace{-14pt}
    \label{fig:egocentric_segmentation}
\end{figure*}

\textbf{Feature Detection \& Matching}: 
Traditional feature detection and extraction techniques, such as SIFT \cite{lowe2004distinctive}, SURF \cite{bay2006surf}, and BRIEF \cite{calonder2010brief}, coupled with RANSAC \cite{fischler1981random} as an outlier rejection technique have been widely adopted in vision-based applications. Within this realm, ORB \cite{mur2015orb} is the first traditional feature-matching method to operate in real-time. Recent development in DL also introduces networks that serve feature-matching purposes, including TILDE \cite{verdie2015tilde}, DeepDesc \cite{simo2015discriminative}, LIFT \cite{yi2016lift}, UCN \cite{choy2016universal}, and SuperPoint \cite{detone2018superpoint} with its variants, such as SuperGlue \cite{sarlin2020superglue} and LightGlue \cite{lindenberger2023lightglue}. 

One closely similar work \cite{tang2021self} proposes a SuperPoint-like framework with DepthNet for depth inference from RGB images to detect 3D keypoints for ego-motion estimation. However, the mentioned work does not maintain the real-time performance of SuperPoint due to the expensive computation of DepthNet. Additionally, 3D keypoint constructed matches from SuperPoint and depth maps generated by DepthNet lead to accumulated errors when encountering out-of-distribution scenes. To alleviate such expensive computations and potential errors, we form a 2D position embedding while training SuperPoint, overlay RGB images with segmentation masks to reduce the search space of the matching procedure, take corresponding depth frames without the need for DepthNet, and obtain 3D correspondences by mapping 2D correspondences produced by the retrained SuperPoint to depth maps that are also from the RGB-D image stream.

\textbf{Point Cloud Alignment \& Registration}: 
Point cloud alignment and registration have predominantly depended on variants of the ICP algorithm, including point-to-point \cite{besl1992method, chen1992object} and point-to-plane \cite{rusinkiewicz2001efficient} strategies. This area of study can be divided into global registration \cite{yang2015go, zhou2016fast}, which primarily operates on 3D correspondence candidates based on point-to-point matches between point clouds, to find the optimal alignment, and local registration \cite{besl1992method, chen1992object, rusinkiewicz2001efficient, park2017colored}, which requires an initial rough alignment to produce a refined alignment that tightly aligns two point clouds. However, global registration often requires multiple iterations to extract good correspondences. Likewise, local registration might lead to inaccurate results or does not converge to an optimal alignment if a rough alignment is not initialized. Therefore, to address the extensive search step for correspondence finding in global registration, we provide the point-set registration procedure with alternative 3D correspondences composed of matched features on RGB images and depth values on corresponding depth maps.

Robust kernels \cite{babin2019analysis, chebrolu2021adaptive} are proposed as an outlier rejection method for 3D correspondences and applied to point cloud alignment problems. Still, these techniques are highly sensitive to hyperparameter choices. To address the sensitivity induced by hyperparameters as in previous works, we utilize kernel density estimation (KDE) -- a non-parametric statistic method to estimate the importance weight of each correspondence, which has recently been used for estimating each point's likelihood \cite{wu2019pointconv}, and as backbones of self-attention modules for density-aware 3D object detection in self-driving cars problems \cite{hu2022point, wang2023net}. Notably, KDE was also mathematically and experimentally proven to be a noise-free and robust registration method for single-object models based on their 3D correspondences \cite{tsin2004correlation} with offline experiments.


%% file: 03_segmentation.tex
\section{Egocentric 3D Object Segmentation} 
\label{sec:segmentation}
In this section, we propose an algorithm to egocentrically segment objects in RGB-D frames, as described in Alg. \ref{alg:create_egocentric_segmentation} with $w$ and $h$ respectively denoted as width and height of RGB images, typically $w = 320$ and $h = 240$. We denote the depth image taken from depth cameras as $\mathcal{D}$, the RGB image taken from the RGB camera as $\mathcal{I}$, and a point cloud with $n$ points as $\mathcal{P} \coloneqq \{\textbf{p}_i\}\text{, for } i = 1, 2, ..., n$, where $\textbf{p}_i = \left[x_i, y_i, z_i\right]^T$ represent points of $\mathcal{P}$ in Cartesian coordinates.
\setlength{\textfloatsep}{1pt}
\begin{algorithm}[t]
    \caption{Egocentric Object Segmentation}
    \label{alg:create_egocentric_segmentation}
    \begin{small}
        \DontPrintSemicolon
        \SetKwInOut{KwIn}{Input}
        \SetKwInOut{KwOut}{Output}
        \SetKwFunction{FMain}{ego\_segment}
        \SetKwProg{Pn}{function}{}{}
        \KwIn{\mbox{$\mathcal{D} \coloneqq$ raw depth image, $\mathcal{I}(w, h) \coloneqq$ raw RGB image}\\
         \mbox{$I^{D}, I^{RGB} \coloneqq$ depth \& RGB cameras intrinsic matrices}\\
         \mbox{$R, T \coloneqq$ extrinsic matrix btw RGB \& depth cameras}}
        \KwOut{\mbox{$\mathcal{P}_{\mathbf{M}} = \{\textbf{p}_i$\} = \{$x_i$, $y_i$, $z_i$\} $\coloneqq$ segmented point cloud}}
        \Pn{\FMain{$\mathcal{D}$, $\mathcal{I}$, $I^{D}$, $I^{RGB}$, $R$, $T$}}{ 
            $\mathcal{D}_{\mathbf{M}}$, $\mathcal{P}_{\mathbf{M}}$ = [ ], [ ]\\
            $\mathbf{M}_{\mathcal{I}} = \texttt{segment}(\mathcal{I})$\\
            \mbox{$\mathcal{D}_{\text{filled}} = \texttt{fill\_holes}(\mathcal{D}, \text{ }\texttt{nearest\_from\_around})$}\\
            $\mathcal{D}_{\text{align}} = \texttt{align\_frame}(\mathcal{D}_{\text{filled}}, \mathcal{I}, I^{D}, I^{RGB}, R, T)$\\
            \For{${u} \in {w} \textbf{ and } {v} \in {h}$}{
                $\mathcal{D}_{\mathbf{M}}(u, v) = \mathcal{D}_{\text{align}}(u, v)$ \textbf{if} $\mathbf{M}_{\mathcal{I}}(u, v) \neq 0$ \textbf{else} 0
            }
            \For{${m} \in {w} \textbf{ and } {n} \in {h}$}{ 
                \If{$\mathcal{D}_{\mathbf{M}}(m, n) \neq 0 $}{
                    $z_{\text{i}} = \mathcal{D}_{\text{rect}}(m, n)$\\
                    $x_{\text{i}},\text{ }y_{\text{i}} = (m - c_{x}^{\text{D}}) \cdot z_{\text{i}} / f_{x}^{\text{D}},\text{ }(n - c_{y}^{\text{D}}) \cdot z_{\text{i}} / f_{y}^{\text{D}}$\\
                    $\mathcal{P}_{\mathbf{M}}\texttt{.append}(\texttt{create\_point}(x_i, y_i, z_i))$
                }
            }
            $\mathcal{P}_{\mathbf{M}} = \texttt{remove\_outliers}(\mathcal{P}_{\mathbf{M}})$\\
            \KwRet{$\mathcal{P}_{\mathbf{M}}$}
        }
    \end{small}
\end{algorithm}

As shown in Fig. \ref{fig:egocentric_segmentation} and Alg. \ref{alg:create_egocentric_segmentation}, the segmentation process first takes $\mathcal{D}$ and $\mathcal{I}$ from the image stream, segments objects of interest presented in $\mathcal{I}$ (Fig. \ref{fig:egocentric_segmentation}a) to obtain object's masks $\mathbf{M}_{\mathcal{I}}$ (Fig. \ref{fig:egocentric_segmentation}b). Next, to guarantee the resulted $\mathcal{P}_{\mathbf{M}}$'s quality, $\mathcal{D}$ is holes-filled using the value from the neighboring pixel closest to the camera, which results in $\mathcal{D}_{\text{filled}}$. Hence, $\mathcal{D}_{\text{filled}}$ is dimensionally aligned with respect to $\mathcal{I}$ (Fig. \ref{fig:egocentric_segmentation}c) with the preprogrammed RGB-D camera's intrinsic and extrinsic parameters, ${I}^{D}$, ${I}^{RGB}$, $R$, and $T$. The aligned depth frame, $\mathcal{D}_{\text{align}}$, is then processed pixel-wise to rectify depth pixels that are outside of $\mathbf{M}_{\mathcal{I}}$ (Fig. \ref{fig:egocentric_segmentation}d) and $\mathcal{D}_{\mathbf{M}}$ is pixel-wise converted into points in $\mathcal{P}_{\mathbf{M}}$ (Fig. \ref{fig:egocentric_segmentation}e) based on optical centers, $(c^D_x, c^D_y)$, and focal lengths, $(f^D_x, f^D_y)$, of depth cameras. Lastly, $\mathcal{P}_{\mathbf{M}}$ is cleaned by removing possible outliers that resulted from holes in depth images in the masks' precision in the earlier step. Note that this procedure also requires skipping a few initial frames (about 5 frames) for the camera to guarantee auto-exposure with brightness control. The computational cost of this procedure on the real robot system is evaluated in Sec. \ref{sec:eval_deploy}. 

\begin{figure}[t]
    \centering
    \includegraphics[width=1.00\linewidth]{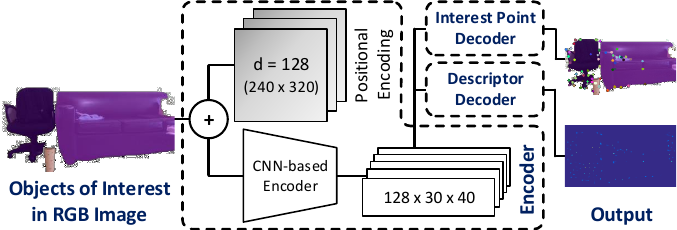}
    \vspace{-13pt}
    \caption{The architecture of the feature-extraction network with 2D-positional embedding, where $\bigoplus$ denotes embedding notation.}
    \label{fig:network_architecture}
\end{figure}

%% file: 04_feature_matching.tex
\section{Feature Detection \& Matching}
\label{sec:feature_matching}

\begin{figure*}[t]
    \centering
    \includegraphics[width=1.00\linewidth]{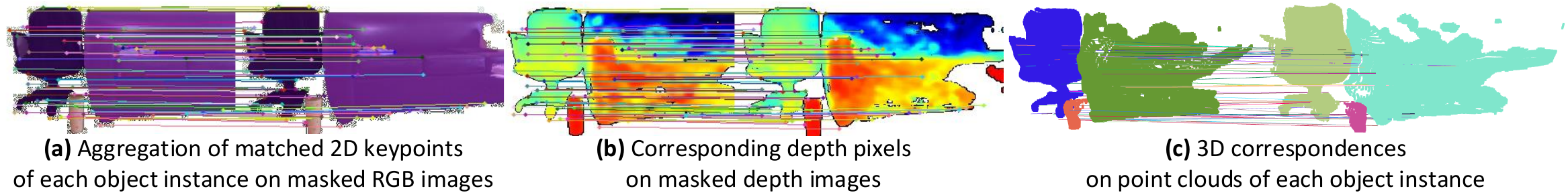}
    \vspace{-15pt}
    \caption{The 3D correspondences matching process includes \textbf{(a)} extracting and matching keypoints between masked RGB images, \textbf{(b)} finding corresponding depth pixels on rectified depth images, and \textbf{(c)} mapping 3D correspondences between point clouds of object instances.}
    \vspace{-18pt}
    \label{fig:finding_3d_corespondences}
\end{figure*}

\subsection{Network Architecture}
Due to the positions of pixels as well as features varying among images in training datasets and in real-world scenes, we apply 1D-positional embedding into the 2D domain to improve the feature-extraction learning process with embedded positional information of every pixel presented on the input RGB image. The formula for a 2D-positional embedding encoder is given below:
\vspace{-4pt}
\begin{equation}
    \small	
    \begin{cases}
      \psi\left(x, y, 2i\right) = \sin \left(\dfrac{x}{D}\right), \qquad \psi\left(x, y, 2j + \dfrac{d}{2}\right) = \sin \left(\dfrac{y}{D}\right)\\
      \psi\left(x, y, 2i + 1\right) = \cos \left(\dfrac{x}{D}\right) \text{, } \psi\left(x, y, 2j + 1 + \dfrac{d}{2}\right) = \cos \left(\dfrac{y}{D}\right)\\
    \end{cases}\
    \vspace{1pt}
    \label{eq:2d_positional_embedding}
\end{equation}
where $\psi$ represents the embedder, $(x, y)$ are pixel coordinates, $D = 10,000^{\text{ }\mathrm{i} \cdot (d/4)^{-1}}$  with $d$ indicating the embedding space's dimension, and $(i, j)$ are integers in $\left[0, d/4 \right)$.

Applying Eq. \ref{eq:2d_positional_embedding} as shown in Fig. \ref{fig:network_architecture}, we design our feature-extraction network with the shared 2D-positional encoder, feed-forwarding to interest point and descriptor decoders. For interest point and descriptor decoders, we inherit these modules from the vanilla SuperPoint \cite{detone2018superpoint}. The training details and the qualitative results of this network design on the HPatches dataset \cite{balntas2017hpatches} are in Sec. \ref{sec:eval_feature_extraction}.

\subsection{Feature Detection \& Matching Strategy}
Since the vanilla approach of SuperPoint takes in entire RGB frames as inputs, it might mismatch a feature on one object in the previous frame to a closely similar feature on another object in the current frame. To eliminate the effect of this, we also leverage the segmented mask in Sec. \ref{sec:segmentation} to provide SuperPoint with masked RGB images of corresponding objects between RGB frames (Fig. \ref{fig:network_architecture}), guaranteeing feature-scanning regions within masked regions. 

As shown in Fig. \ref{fig:finding_3d_corespondences}, we first create masked RGB images of each corresponding object in two consecutive frames and then apply the retrained SuperPoint on each pair of images (\textit{e.g.,} cups, couches, and chairs as illustrated) to extract and match 2D keypoints within each object instance. This avoids feature mismatching across non-corresponding objects. We then aggregate matched features from separated objects in the previous step (Fig. \ref{fig:finding_3d_corespondences}a), and find the corresponding matched pixels in the aligned depth images (Fig. \ref{fig:finding_3d_corespondences}b). With the image coordinates in Fig. \ref{fig:finding_3d_corespondences}a and depth values from Fig. \ref{fig:finding_3d_corespondences}b, we compute 3D correspondences between point clouds using \texttt{line 11} in Alg. \ref{alg:create_egocentric_segmentation}. The correspondences between point clouds of each object instance presented in the scene are shown in Fig. \ref{fig:finding_3d_corespondences}c. 

%% file: 05_reconstruction.tex
\section{Point Cloud Alignment \& Registration}
\label{sec:scene_reconstruction}
In this section, we elaborate on the point cloud alignment procedure with extracted 3D correspondences between $\mathcal{P}_{t}$ and $\mathcal{P}_{t-1}$ from Sec. \ref{sec:feature_matching}. Without loss of generality, we denote 3D corresponding point sets in the current view and previous view as $\mathcal{K}_{t}$ and $\mathcal{K}_{t-1}$, respectively, and each 3D pointset of $m$ points $\mathcal{K}_{i} \coloneqq \{ \textbf{p}^{\mathcal{K}_{i}}_{j} \}$, for $j = 1, 2, ..., m$, where $ \textbf{p}^{\mathcal{K}_{i}}$ are points of $\mathcal{K}$ in Cartesian coordinates, with $m \ll n$.

\subsection{3D Correspondence Importance Weighting}
\label{sec:importance_weighting}
Noticing that the feature detection process (Sec. \ref{sec:feature_matching}) mainly concentrates on segmented regions on RGB images from Sec. \ref{sec:segmentation}, these procedures produce $\mathcal{K}_{t}$ (or $\mathcal{K}_{t-1}$) that are bounded in several regions in 3D space, which conceptually is a mixture of distributions with dense regions, which are dense with points, and outliers with respect to the spatial distribution of $\mathcal{K}_{t}$. 

\subsubsection{Weight Initialization}
\label{sec:scene_reconstruction_weight_init}
To weigh each correspondence in $\mathcal{K}_{t}$ and $\mathcal{K}_{t-1}$, we first initialize that point's weight via its number of neighboring points inside a specific radius and then compute the cumulative density of points within a region in a dimension-wise sense to obtain the weight of that point. 

\subsubsection{Density Estimation}
The density of the unknown distribution of $\mathcal{K}_{t}$ is estimated using (i) KDE and (ii) Improved Sheather-Jones (ISJ) algorithm for data-driven optimal bandwidth selection \cite{wand1994fast} since we (i) could not make any assumptions toward the distribution of $\mathcal{K}_{t}$ and (ii) would like our transformation in Sec. \ref{sec:point_cloud_alignment} to be robust with outliers, which are less-relevant points in $\mathcal{K}_{t}$.

This procedure is illustrated in Alg. \ref{alg:weighting_points}. We first initialize the weighing coefficient, $w_{i}$, for each $\textbf{p}^{\mathcal{K}_{t}}_{i}$, as the number of neighboring points, $|\mathcal{N}_{i}|$, within a radius, $r$. Next, to avoid the curse of dimensionality when applying non-parametric methods and since $x$, $y$, and $z$ of each $\textbf{p}^{\mathcal{K}_{t}}_{i}$ are independent of each other in our problem, we treat them as three random variable vectors $X$, $Y$, and $Z$. Taking $X$ as an illustration, we apply KDE for each random-variable vector as follows:
\vspace{-4pt}
\begin{equation}
    \hat{f}_X(x) = \frac{1}{mH}\sum^{m}_{i=1} w_{i} \cdot K\left[\frac{x - x_{i}}{H}\right]
    \vspace{-4pt}
    \label{eq:kde}
\end{equation}
where $H$ represents the optimal bandwidth for the kernel, $K$, which is often chosen as the Gaussian distribution.

\setlength{\textfloatsep}{1pt}
\begin{algorithm}[t]
    \caption{Important Points Weighting}
    \label{alg:weighting_points}
    \begin{small}
        \DontPrintSemicolon
        \SetKwInOut{KwIn}{Input}
        \SetKwInOut{KwOut}{Output}
        \SetKwFunction{FMain}{weigh\_corresponding\_points}
        \SetKwProg{Pn}{function}{}{}
        \KwIn{$\mathcal{K}_{t} \coloneqq$ set of 3D corresponding points of $\mathcal{P}_{t}$\\
        $r \coloneqq$ radius of sphere around a 3D point}
        \KwOut{\mbox{$W =\text{diag}(w_{1}, ..., w_{n}) \coloneqq$ diagonal weighing matrix}}
        \Pn{\FMain{$\mathcal{K}_{t}$, $r$}}{ 
            $w$, $F$ = [ ], [ ]\\
            \mbox{$w\texttt{.append}\left(|\mathcal{N}_{i}|\text{ }= \texttt{neighbor}(\textbf{p}^{\mathcal{K}_{t}}_{i}, r)\right)$ \textbf{for} $\textbf{p}^{\mathcal{K}_{t}}_{i} \in \mathcal{K}_{t}$}\\
            \For{$k \in \left\{\mathcal{K}_{t}[:,0], \mathcal{K}_{t}[:,1], \mathcal{K}_{t}[:,2]\right\}$}{
                \mbox{$F\texttt{.append}(\texttt{fft\_KDE}(k, w, H = \texttt{optband\_ISJ}(k)))$}
            }
            $w = w \cdot f_{i}$ \textbf{for} $f_{i} \in F$\\
            \KwRet{$\texttt{diag}\left(w^{\text{T}}\right)$}
        }
    \end{small}
\end{algorithm}

To reduce computational complexity of $\mathcal{O}(m^2)$ from Eq. \ref{eq:kde}, the binning (alternative) approximation for Eq. \ref{eq:kde} is introduced \cite{wand1994fast}. Written in terms of grid points $\{\textbf{g}_{j}\} \in \left[\textbf{g}_{1}, \textbf{g}_{M}\right]$ covering all $\textbf{p}^{\mathcal{K}_{t}}_{i}$, and grid counts $\{c_{j}\}$ designating weights chosen to represent the number of $\textbf{p}^{\mathcal{K}_{t}}_{i}$'s that are near $\textbf{g}_{j}$, for $j = 1, 2, ..., M$, and $M \neq m$, yields:
\vspace{-5pt}
\begin{equation}
    \small
    \widetilde{f_{X, \textbf{g}_j}} \coloneqq \tilde{f}_X(g_{j}) = \frac{1}{mH}\sum^{M}_{i=1} w_{i} c_{i} \cdot K\left[\frac{\textbf{g}_{j} - \textbf{g}_{i}}{H}\right] \approx \hat{f}_X(x)
    \vspace{-5pt}
    \label{eq:binning_kde}
\end{equation}

\begin{figure*}[t]
    \centering
    \includegraphics[width=1.00\linewidth]{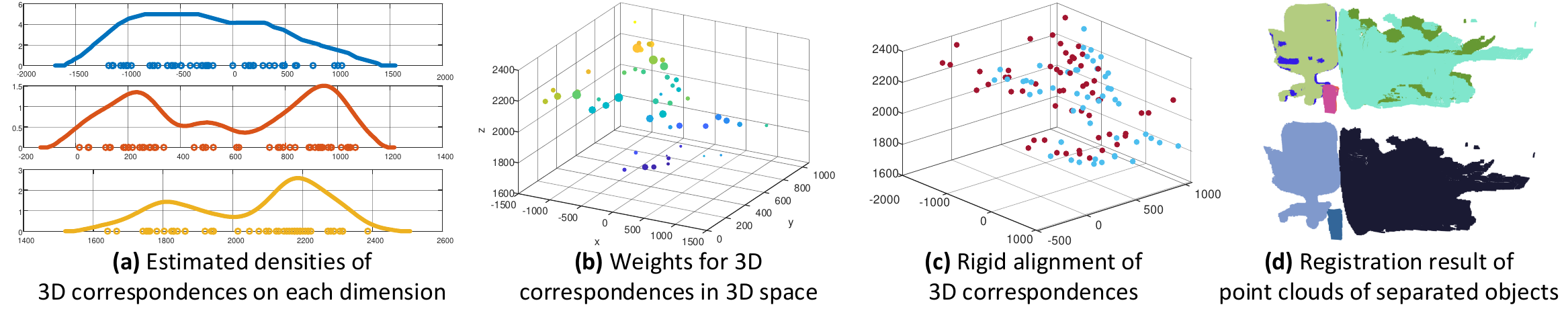}
    \vspace{-16pt}
    \caption{The point cloud alignment process includes \textbf{(a)} dimension-wise estimating densities of 3D correspondences along x-axis (blue), y-axis (orange), and z-axis (yellow), \textbf{(b)} computing weights for 3D correspondences, \textbf{(c)} solving for the optimal rigid transformation based on 3D correspondences and their weights, and \textbf{(d)} aligning point clouds (top) with re-colorization based on each object instance (down).}
    \vspace{-17pt}
    \label{fig:3d_reconstruction}
\end{figure*}

Taking advantage of the FFT's time complexity of $\mathcal{O}(M\log M)$, Eq. \ref{eq:binning_kde} is then rewritten in the form of convolution as:
\vspace{-4pt}
\begin{equation}
    \small
    \widetilde{f_{X, \textbf{g}_j}} = \sum^{M-1}_{i=-(M-1)} c_{j-i} \textbf{k}_{i} \text{ with } \textbf{k}_{i} = \frac{w_{i}}{m} \cdot K\left[\frac{(\textbf{g}_{M} - \textbf{g}_{1}) i}{H}\right]
    \vspace{-5pt}
    \label{eq:fft_kde}
\end{equation}

Together with the binning process in Eq. \ref{eq:binning_kde}, the time complexity of Alg. \ref{alg:weighting_points} is $\mathcal{O}(m + M\log M)$. The optimal bandwidth, $H$, is obtained from the ISJ algorithm with $\mathcal{K}_t$. 

\subsubsection{Weight Updating}
Applying Eq. \ref{eq:fft_kde}, we obtain density functions $\hat{f}_X$, $\hat{f}_Y$, and $\hat{f}_Z$ for $X$, $Y$, and $Z$, respectively, as illustrated in Fig. \ref{fig:3d_reconstruction}a. Hence, $w_{i}$ for each $\textbf{p}^{\mathcal{K}_{t}}_{i} \coloneqq \{x_{i}, y_{i}, z_{i}\}$ is updated from the naive estimation in Sec. \ref{sec:scene_reconstruction_weight_init}:
\vspace{-4pt}
\begin{equation}
    w_{i} \coloneqq w_{i} \cdot \hat{f}_{X}(x_{i}) \cdot \hat{f}_{Y}(y_{i}) \cdot \hat{f}_{Z}(z_{i})
    \vspace{-3pt}
    \label{eq:weight_estimation}
\end{equation}

Using Eq. \ref{eq:weight_estimation} for all points in $\mathcal{K}_t$, we obtain the weighing $m \times m$ diagonal matrix, $W = \text{diag}(w_{1}, ..., w_{m})$, The weights are illustrated in Fig. \ref{fig:3d_reconstruction}b, with bigger points portraying larger weights, and vice versa.

\subsection{Rigid Motion for Point Cloud Alignment}
\label{sec:point_cloud_alignment}
To find a rigid transformation matrix, $T$, for $\mathcal{P}_{t}$ and $\mathcal{P}_{t-1}$, that aligns $\mathcal{K}_{t}$ and $\mathcal{K}_{t-1}$ in $\mathbb{R}^3$, we define the optimizer in terms of a rotation matrix, $R$, a translation vector, $\textbf{t}$, and the weighing coefficient, $w_i$ for $\textbf{p}^{\mathcal{K}_{t}}_{i}$ and $\textbf{p}^{\mathcal{K}_{t-1}}_{i}$, as follows:
\vspace{-4pt}
\begin{equation}
    T = [R, \textbf{t}] = \argminE_{R,\text{ }\textbf{t}} \sum^{n}_{i = 1} w_{i} \left|\left|\textbf{p}^{\mathcal{K}_{t}}_{i} - (R\textbf{p}^{\mathcal{K}_{t-1}}_{i} + \textbf{t}) \right|\right|^2
    \vspace{-4pt}
    \label{eq:transformation_optimizer}
\end{equation}

From Eq. \ref{eq:transformation_optimizer}, the optimal $3 \times 1$ translation vector, \textbf{t}, is:
\vspace{-3pt}
\begin{equation}
    \textbf{t} = \overline{\textbf{p}^{\mathcal{K}_{t}}} - R \cdot \overline{\textbf{p}^{\mathcal{K}_{t-1}}}
    \vspace{-5pt}
    \label{eq:optimal_translation}
\end{equation}
\vspace{-7pt}
\begin{equation*}
    \text{where } \overline{\textbf{p}^{\mathcal{K}_{t}}} \coloneqq \frac{\sum_{i=1}^{n} w_{i} \textbf{p}^{\mathcal{K}_{t}}_{i}}{\sum_{i=1}^{n} w_{i}} \text{ and } \overline{\textbf{p}^{\mathcal{K}_{t-1}}} \coloneqq \frac{\sum_{i=1}^{n} w_{i} \textbf{p}^{\mathcal{K}_{t-1}}_{i}}{\sum_{i=1}^{n} w_{i}} \text{}
    \vspace{-1pt}
\end{equation*}
are weighted centroids of $\mathcal{K}_{t}$ and $\mathcal{K}_{t-1}$, respectively.

Meanwhile, to find the optimal $3 \times 3$ rotation matrix, $R$, we first compute the covariance matrix, $C$, which is the product of the $3 \times m$ matrix, $D_{t}$, column-wise containing deviations of $\textbf{p}^{\mathcal{K}_{t}}_{i}$ from $\overline{\textbf{p}^{\mathcal{K}_{t}}}$ in $\mathcal{K}_{t}$ and the $3 \times m$ matrix, $D_{t-1}$, column-wise containing deviations of $\textbf{p}^{\mathcal{K}_{t-1}}_{i}$ from $\overline{\textbf{p}^{\mathcal{K}_{t-1}}}$ in $\mathcal{K}_{t-1}$, and $W$ obtained from Alg. \ref{alg:weighting_points} in Sec. \ref{sec:importance_weighting}:
\vspace{-4pt}
\begin{equation}
    C = D_{t} W D_{t-1}^{\text{T}}
    \vspace{-3pt}
    \label{eq:covariance_matrix}
\end{equation}

Hence, the optimal rotation matrix is computed in terms of rotation matrices, $U$ and $V$, resulting from the singular value decomposition $U\Sigma V^{\text{T}}$ of $C$ from Eq. \ref{eq:covariance_matrix}, as follows: 
\vspace{-4pt}
\begin{equation}
    \small
    R = V \begin{bmatrix}
        1 & & &\\
        & & \ddots &\\
        & & & \text{det}(VU^{\text{T}})
    \end{bmatrix} U^{\text{T}}
    \vspace{-3pt}
    \label{eq:optimal_rotation}
\end{equation}

Note that the scaling matrix, $\Sigma$, is omitted in Eq. \ref{eq:optimal_rotation} as we need to preserve object sizes and shapes in $\mathcal{P}_{t-1}$ in Eq. \ref{eq:aligning_clouds}.

\begin{figure}[t]
    \centering
    \includegraphics[width=1.00\linewidth]{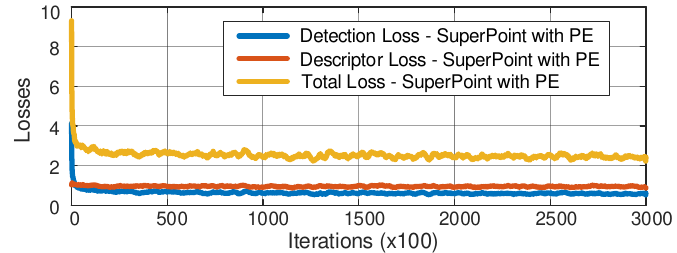}
    \vspace{-15pt}
    \caption{Training performance of SuperPoint with positional embedding, including detection loss, descriptor loss, and total loss.}
    \vspace{-7pt}
    \label{fig:superpoint_performance}
\end{figure}

\begin{figure}[t]
    \centering
    \includegraphics[width=1.00\linewidth]{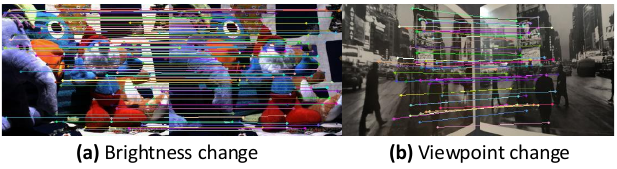}
    \vspace{-17pt}
    \caption{Qualitative results of Superpoint with positional embedding with brightness and viewpoint change images from HPatches.}
    \vspace{-2pt}
    \label{fig:hpatches_quality}
\end{figure}

Therefore, from Eq. \ref{eq:optimal_translation} and Eq. \ref{eq:optimal_rotation}, the $4 \times 4$ homogeneous rigid transformation matrix solving Eq. \ref{eq:transformation_optimizer} for
$\mathcal{K}_{t}$ and $\mathcal{K}_{t-1}$ and eventually for $\mathcal{P}_{t}$ and $\mathcal{P}_{t-1}$, is defined as:
\vspace{-5pt}
\begin{equation}
    T = \begin{bmatrix}
        R & \textbf{t}\\
        0_{1, 3} & 1
    \end{bmatrix}
    \vspace{-3pt}
    \label{eq:rigid_homogeneous_transformation}
\end{equation}

With the homogeneous transformation, $T$, from Eq. \ref{eq:rigid_homogeneous_transformation}, we align two multiview point clouds, $\mathcal{P}_{t}$ and $\mathcal{P}_{t-1}$, as follows:
\vspace{-5pt}
\begin{equation}
    \mathcal{P}_{\text{align}} = \mathcal{P}_{t} + \left[R \textbf{p}^{\mathcal{P}_{t-1}}_{i} + \textbf{t}\right]_{i = 1}^{n} \coloneqq \mathcal{P}_{t} + T\left(\mathcal{P}_{t-1}\right)
    \label{eq:aligning_clouds}
\end{equation}

%% file: 06_evaluation.tex
\section{Evaluation \& Experimental Results}
\label{sec:eval}

\subsection{Performance of SuperPoint with Positional Embedding}
\label{sec:eval_feature_extraction}
We retrain SuperPoint with 2D positional embedding with $d = 128$ (Sec. \ref{sec:feature_matching}) on the MS COCO 2014 dataset with interest points labeled by MagicPoint, which is refined on a synthetic auto-generated dataset (Fig. \ref{fig:superpoint_performance}). All images are resized to $240 \times 320$ and are augmented by adding random brightness and contrast, Gaussian noises, shades, and motion blur. The training process takes place on the NVIDIA RTX 4090 GPU with 10 epochs (300,000 iterations).

Also, Fig. \ref{fig:hpatches_quality} shows qualitative results of SuperPoint with positional embedding on images from HPatches \cite{balntas2017hpatches}, illustrating the robustness in the two most common scenarios: brightness change (Fig. \ref{fig:hpatches_quality}a) and viewpoint change (Fig. \ref{fig:hpatches_quality}b).

\subsection{Point Cloud Alignment Error at Multiple Angles}
\label{sec:eval_pc_alignment}
We evaluate the performance of our point cloud alignment (Sec. \ref{sec:point_cloud_alignment}) by calculating Root Mean Squared Error (RMSE) between the two correspondence sets, $\mathcal{K}_{t-1}$ and $\mathcal{K}_{t}$. To do this, we move the camera to various angles with a 2-meter distance from the scene on a planar surface, including $0^{\circ}$ (stands at its initial position), $\pm 10^{\circ}$, $\pm 20^{\circ}$, $\pm 30^{\circ}$, and $\pm 45^{\circ}$, and computing their alignment using RMSE. 

\begin{figure}[t]
    \centering
    \includegraphics[width=1.00\linewidth]{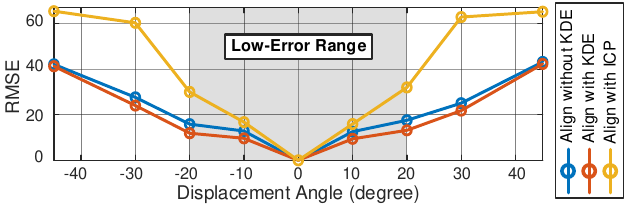}
    \vspace{-15pt}
    \caption{Alignment error computed in RMSE at multiple angles of alignments with/without using KDE, and point-to-point ICP.}
    \vspace{-2pt}
    \label{fig:rmse}
\end{figure}

As shown in Fig. \ref{fig:rmse}, RMSE is larger as the displacement angle gets wider. The results also prove that KDE is effectively involved in the process of reducing alignment errors. Meanwhile, ICP registration underperforms our alignment method, as the rough alignment assumption is not guaranteed when capturing views at a wide displacement angle.

\subsection{Deployment on Baxter Robot}
\label{sec:eval_deploy}
\subsubsection{Experimental Setup}
To obtain RGB-D images at multiple views, we mount the Intel RealSense D435i RGB-D camera on the Baxter robot with a setup scene of a table, a chair, a bag, and two plastic cups on the table. Prior to the experiment, the robot is pre-calibrated using the multiplanar self-calibration method \cite{dang2023multiplanar} for precise object manipulation.

\begin{figure*}[t]
    \centering
    \includegraphics[width=1.00\linewidth]{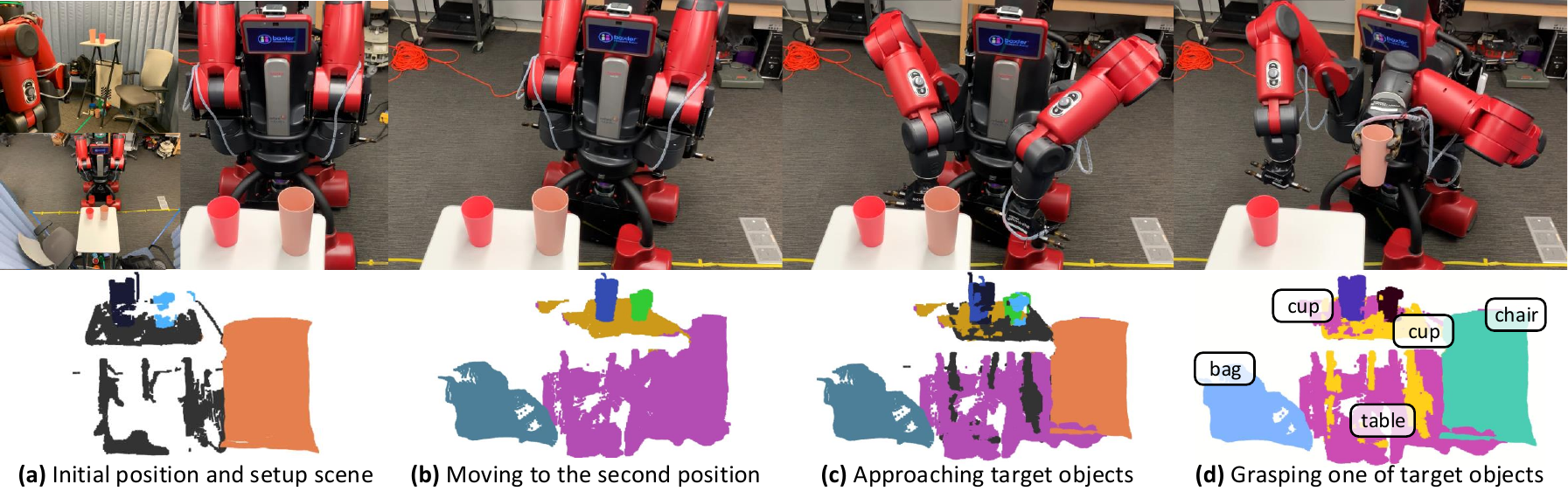}
    \vspace{-14pt}
    \caption{Experiment setup with (top) the Baxter \textbf{(a)} observing the scene at its first view, \textbf{(b)} moving to and capturing the second scene of 20-degree displacement, \textbf{(c)} approaching the target objects, and \textbf{(d)} grasping one of the cups; and (down) each-step results in the pipeline.}
    \vspace{-18pt}
    \label{fig:experiment_setup}
\end{figure*}

\subsubsection{Robot Moving \& Capturing Scene at Multiple Views}
To take multiple views of the scene, the Baxter first stands in one position, takes one view (Fig. \ref{fig:experiment_setup}a), moves to another angle, and takes another view (Fig. \ref{fig:experiment_setup}b), with the robot's movements supported by the Dataspeed mobility base. 

Specifically, the process takes place on two separate computers, including one acquiring data from the RGB-D camera and the main computer, with synchronization using the Robot Operating System (ROS) messages through TCP/IP protocols between them. At each position, once the computer in charge of the robot's vision receives a motion command from the main computer, it executes motion planning and notifies the main computer when the motion command is done. Then the main computer captures a point cloud from the RGB-D camera. The process is then repeated for other angles, which provides the knowledge of the scene from multiple views.

\subsubsection{Segmenting \& Aligning Multiview Point Clouds}
After capturing multiview point clouds, the Baxter first egocentrically segment objects in the scene (Sec. \ref{sec:segmentation}), match 3D correspondences between two views (Sec. \ref{sec:feature_matching}), solves for the transformation of weighted 3D correspondences for rigid alignment (Sec. \ref{sec:scene_reconstruction}), and eventually gives itself a scene understanding of the bag in the second view, the chair in the first view, and the table and the cups in both views.

\subsubsection{Approaching \& Grasping Target Objects}
To demonstrate the feasibility of robotic grasping with 3D semantic scene perception for cobot-like systems, we navigate the Baxter to approach the target objects in the scene. When the target objects are within the robot's workspace (Fig. \ref{fig:experiment_setup}c), the Baxter can grasp those objects efficiently (Fig. \ref{fig:experiment_setup}d).

\vspace{-4pt}
\begin{table}[h]
    \centering
    \setlength\tabcolsep{1pt}
    \begin{tabular}{|c |c |c |c |c |c |} 
        \hline \textbf{ Task } & \textbf{ Device } & \textbf{ Time Complexity } & \textbf{ Runtime }\text{ }\\
        \hline \text{ }Egocentric Segmentation\text{ } & \text{ }GPU\text{ } & n/a & \text{ }96.49 ms\text{ } \\
        \hline \text{ }Keypoint Extraction\text{ } & \text{ }CPU\text{ } & n/a & 158.91 ms\\ 
        \hline \text{ }Keypoint Matching\text{ } & \text{ }CPU\text{ } & n/a & 2.35 ms\\ 
        \hline \text{ }Keypoint Weighting\text{ } & \text{ }CPU\text{ } & $\mathcal{O}(m+M\log M)$ & 2.07 ms\\ 
        \hline \text{ }Point Cloud Alignment\text{ } & \text{ }CPU\text{ } & $\mathcal{O}(m^{3})$ & 0.73 ms\\ 
        \hline
    \end{tabular}
    \vspace{-3pt}
    \caption{Time complexity and runtime in milliseconds (ms) of segmentation, keypoint extraction and matching, keypoint weighting, and point cloud alignment, on onboard processing units.}
    \label{table:runtime}
    \vspace{-8pt}
\end{table}

\subsubsection{Time Complexity on Conventional Hardware}
We use the OpenVINO library \cite{opevino} to deploy YOLOv8n on the native onboard GPU (Intel HD Graphics 4000) of the Intel NUC5i7RYH PC -- the computer processing RGB-D images -- on our customized software framework \cite{dang2023perfc}. The average runtime across onboard devices is reported in Table \ref{table:runtime}. 

\subsection{Demonstration}
The video demonstrates a scenario with the Baxter (1) standing in one position, (2) moving to a position of $20^{\circ{}}$ displacement while (3) performing the pipeline, and (4) grasping the plastic cup: https://youtu.be/-dho7l\_r56U.

%% file: 07_limitations.tex
\section{Limitations \& Future Works}
\label{sec:limitations}
\begin{figure}[t]
    \centering
    \includegraphics[width=1.00\linewidth]{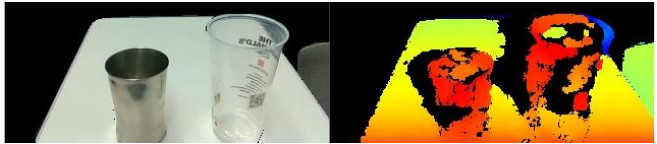}
    \vspace{-15pt}
    \caption{Failure of depth perception of transparent and shiny cups.}
    \label{fig:failure_case}
\end{figure}

As we specifically aim for indoor robots with RGB-D cameras, our primary limitation is its inapplicability to far-field omnidirectional perception (\textit{e.g.}, LiDAR-based perception). In addition, the depth camera also fails to recognize transparent and shiny objects, as illustrated in Fig. \ref{fig:failure_case}, which makes the robot grasping inaccurate. Therefore, we reserve the tasks of perceiving transparent/shiny objects and aligning RGB-D cameras and LiDAR sensors for future work. 

%% file: 08_conclusions.tex
\section{Conclusions}
This work introduces an end-to-end pipeline with instance segmentation, feature matching, and alignment for mobile cobots with RGB-D perception. Our pipeline first segments objects of interest in the scene as the robot moves and matches features between consecutive RGB images before obtaining 3D correspondences via depth maps. The 3D correspondences are then statistically weighted based on their spatial distribution using KDE for rigid point cloud alignment. Through real-world experiments on the 7-DOF dual-arm Baxter robot with an Intel RealSense D435i RGB-D camera, the results show the robot is able to semantically understand the setup scene and grasp the target objects. 

%% file: 09_acknowledgement.tex
\section{Acknowledgment}
We thank Nghia Le (University of Toronto) for fruitful discussions on selecting non-parametric statistical methods.